\documentclass[cameraready]{Interspeech}

\title{Spectral Boundary Failures in Voiceless Obstruents Across Modern Neural TTS Architectures}

\author[affiliation={1}]{Sriyugesh}{Bhyravajjula}
\author[affiliation={2,3}]{Kevin}{Tang}
\author[affiliation={1}]{Ayushi}{Pandey}

\address{
    $^1$ Independent Researcher, India \\
    $^2$ Department of English Language and Linguistics, Institute of English and American Studies, Faculty of Arts and Humanities, Heinrich Heine University Düsseldorf, Germany \\
    $^3$ Department of Linguistics, University of Florida, United States of America
}

\email{bsriyugesh@gmail.com}


\usepackage{comment}
\usepackage{subcaption}
\usepackage{graphicx}
\usepackage{tipa}
\usepackage{pgfplots}
\usepgfplotslibrary{groupplots}
\usepackage{siunitx}
\pgfplotsset{compat=1.18}

\definecolor{clrDiffGAN}{RGB}{31,119,180}
\definecolor{clrFS2}{RGB}{255,127,14}
\definecolor{clrGlow}{RGB}{44,160,44}
\definecolor{clrGrad}{RGB}{214,39,40}
\definecolor{clrPFlow}{RGB}{148,103,189}
\definecolor{clrTaco}{RGB}{140,86,75}
\definecolor{clrTrans}{RGB}{227,119,194}
\definecolor{clrVITS}{RGB}{127,127,127}

\begin{document}

\maketitle
{\renewcommand{\thefootnote}{}\footnotetext{Submitted to Interspeech 2026.}}

\begin{abstract}

Prior work has shown that voiceless fricatives exhibit statistically significant acoustic-phonetic deviations from natural speech across several TTS architectures.
Whether these deviations generalize across broader paradigms, vary with phonetic context, and surface in perceptual representations remains open.

We evaluate voiceless fricatives and stops at CV and VC boundaries across eight TTS systems spanning four architectural classes, using acoustic-phonetic features and RASTA-PLP cepstral coefficients. Deviations are consistent across all systems, confirming the trend is not architecture-specific. CV and VC boundaries produce asymmetric profiles, indicating different boundary-handling failures at each transition
direction. Both feature sets recover the same patterns, suggesting the distortions are perceptually salient. Architecture-specific signatures link distinct deviation profiles to alignment strategy and other architectural nuances. 

\end{abstract}

\section{Introduction}
Rapid advances in modern text-to-speech (TTS) have enabled highly natural synthetic speech \cite{kaur_conventional_2023}. However, overall naturalness ratings may obscure fine-grained segmental deviations that remain perceptually relevant \cite{pandey2023listener}. Although segmental evaluations of TTS systems remain relatively limited, existing studies have consistently shown that voiceless obstruents exhibit significant acoustic-phonetic deviations from natural speech \cite{Pandey2022ProductionCO}, a finding subsequently corroborated on modern neural TTS architectures \cite{bhyravajulla25_ssw}.
In the latter study, end-to-end systems showed these deviations most prominently, while architectures employing flow- and diffusion-based acoustic modeling achieved comparatively smaller deviations. It remains unclear, however, whether these differences stem from architectural properties, or whether voiceless obstruents are intrinsically harder to model regardless of paradigm.

Supporting the perceptual relevance of this line of work, \cite{pandey2023listener} found that listeners rated utterances rich in obstruents as more machine-like compared to sonorant-rich utterances selected at random. Taken together with the acoustic findings of \cite{bhyravajulla25_ssw} and \cite{Pandey2022ProductionCO}, which identify voiceless fricatives as the segment class showing the most pronounced deviations from
the natural speech baseline, this motivates a closer examination of obstruents in modern TTS. The study on modern TTS systems further noted that deviations were more pronounced on sentences withheld from training, a condition we adopt here to maintain consistency with that setting.

We extend this line of inquiry in three directions. First, we expand architectural coverage to eight TTS systems, all trained on LJSpeech to remain comparable with
prior work, spanning autoregressive, non-autoregressive, flow-based, and diffusion-based acoustic modeling approaches. This breadth allows us to assess whether deviation trends are universal or tied to specific modeling strategies.
Second, prior work hypothesized that phonetic context may modulate the observed deviations; we investigate this directly by analyzing voiceless obstruents at
consonant--vowel (CV) and vowel--consonant (VC) boundaries. Third, we complement acoustic-phonetic features with RASTA-PLP cepstral analysis to examine whether the same trends surface in a perceptually motivated representation.

Our study addresses the following research questions: (\textbf{RQ1})~Do deviations in voiceless obstruents persist across an expanded set of architectures? (\textbf{RQ2})~Does phonetic context
reveal deeper or directionally asymmetric patterns of deviation at CV and VC boundaries? (\textbf{RQ3})~Do architecture-specific deviation profiles emerge that reflect differences in modeling architectures?
\textbf{Expectations:} We expect voiceless obstruents to continue showing significant deviations, consistent with prior evidence. We further hypothesize that CV and VC boundaries will produce
asymmetric profiles, given the qualitatively different spectral transition each involves: from aperiodic turbulence noise into periodic formant structure, and vice versa. Finally, we anticipate that the expanded architecture set will surface
paradigm-specific signatures that illuminate the robustness and limitations of current acoustic modeling approaches.

The paper is structured as follows. Section~\ref{sec:expsetsup} describes the experimental setup, including dataset, architectures, alignment procedure, and feature extraction.
Section~\ref{sec:results} presents quantitative results for both acoustic-phonetic and RASTA-PLP features. Section~\ref{sec:discussion} offers a synthesis of the findings, linking observed deviation
patterns to architectural hypotheses. Section~\ref{sec:conclusion} concludes.  All code for
replicating this work will be made publicly available.

\section{Experimental Setup}\label{sec:expsetsup}

\begin{table*}[t]
\centering
\caption{Summary of evaluated TTS architectures. Properties are limited to those directly relevant to the analysis in Sections~3 and~4.}
\label{tab:architectures}
\setlength{\tabcolsep}{5pt}
\resizebox{\linewidth}{!}{
\begin{tabular}{lllllll}
\toprule
\textbf{System} & \textbf{Class} & \textbf{Generation} & \textbf{Alignment} & \textbf{Training Loss} & \textbf{Variance Adaptor} & \textbf{Waveform} \\
\midrule
Tacotron2 \cite{shen2018naturalttssynthesisconditioning}       & Autoregressive        & Sequential  & Soft attention (location-sensitive) & MSE        & --      & 2-stage (HiFi-GAN) \\
Transformer TTS \cite{li2019neuralspeechsynthesistransformer} & Autoregressive        & Sequential  & Soft attention (cross-attention)     & MSE        & --      & 2-stage (HiFi-GAN) \\
FastSpeech2 \cite{ren2022fastspeech2fasthighquality}     & Non-autoregressive    & Parallel    & Hard (forced aligner)               & MSE        & \checkmark (dur., pitch, energy) & 2-stage (HiFi-GAN) \\
GlowTTS \cite{kim2020glowttsgenerativeflowtexttospeech}         & Non-autoregressive    & Parallel    & MAS                                 & Flow (NLL) & --      & 2-stage (HiFi-GAN) \\
GradTTS \cite{popov2021gradttsdiffusionprobabilisticmodel}        & Non-autoregressive    & Iterative   & MAS                                 & Diffusion  & --      & 2-stage (HiFi-GAN) \\
PFlowTTS \cite{kim2023pflow}       & Non-autoregressive    & Parallel    & MAS                                 & Flow (NLL) & --      & 2-stage (HiFi-GAN) \\
DiffGAN-TTS \cite{liu2022diffganttshighfidelityefficienttexttospeech}     & Non-autoregressive    & Iterative   & Hard (FS2 prior)                    & GAN + diff.& \checkmark (inherited) & 2-stage (HiFi-GAN) \\
VITS \cite{kim2021conditionalvariationalautoencoderadversarial}           & End-to-end            & Parallel    & MAS (stochastic dur.)               & VAE + GAN  & --      & End-to-end \\
\bottomrule
\end{tabular}
}
\end{table*}

\subsection{Dataset Description}
All TTS systems were trained on LJSpeech \cite{ljspeech17}, a standard benchmark for TTS evaluation due to its high recording quality and phonetically diverse content. It is a single-speaker corpus consisting of 13,100 short audio clips totaling approximately 23 hours and 55 minutes of read speech from seven non-fiction books. Audio is stored as 16-bit PCM WAV files at 22,050 Hz.

\subsection{Experimental conditions}
Phoneme boundaries were obtained using the Montreal Forced Aligner (MFA)~\cite{mcauliffe2017montreal} with its default US English ARPA acoustic model and pronunciation dictionary. Segments were classified by phoneme identity and extracted at their CV and VC boundaries. All acoustic-phonetic and RASTA-PLP \cite{225957} features were extracted in Python.  The distribution of obstruents across CV and VC contexts and contexts themselves was verified to be balanced within each system, supporting direct comparison between boundary conditions.

Synthesis was evaluated on the held-out test sentences of each architecture, as specified in their respective official repositories. The test sets are not identical across systems, and no sub-sampling was performed in order to preserve reproducibility with each system's original evaluation conditions. Crucially, each system is compared exclusively against the natural speech ground truth derived from its own test set. Comparisons are therefore always within-system; the difference in test set size across architectures does not affect the validity of any individual system's results or the
cross-system trends reported here. 

\subsection{Feature extraction}

A brief description of the TTS systems is provided in Table~\ref{tab:architectures}. All speech was synthesized using publicly available implementations of the above models with default hyperparameters trained on LJSpeech. Two feature categories were extracted per segment. \textbf{Acoustic-phonetic features}: seven measures adopted from prior work~\cite{bhyravajulla25_ssw} for their established
discriminative properties for voiceless obstruents -- duration, dynamic amplitude, peak amplitude, peak frequency, RMS amplitude, spectral shape, and spectral tilt. \textbf{RASTA-PLP (RPLP) cepstral  coefficients}: computed from 16\,kHz (it is worth noting that both synthetic and human speech have been downsampled to 16\,kHz to maintain consistency) audio using a 25\,ms Hann window, 10\,ms hop, 21 Bark-spaced critical bands, first-order IIR RASTA filter ($\alpha = 0.97$), equal-loudness weighting, cube-root compression, and 12th-order LPC, yielding 13 coefficients per frame averaged across the segment. RASTA-PLP was preferred over MFCCs for its auditory frequency warping and temporal band-pass filtering, which suppress static noise while amplifying spectral transitions at phonetic boundaries \cite{225957}.
Coefficients C$_0$--C$_{12}$ were screened via the Kruskal-Wallis H-test; C$_0$--C$_4$ were retained on the basis of significance ($p < 0.001$) and interpretability. Their acoustic-perceptual interpretations are as follows: C$_0$ reflects overall log-energy as perceived by the auditory system, owing to equal-loudness pre-emphasis and cube-root compression. C$_1$ captures broad spectral balance, approximating the energy distribution between the lower and upper halves of the frequency range. C$_2$ encodes second-order spectral curvature, approximating spectral tilt across Bark-warped frequency bands. C$_3$ and C$_4$ measure spectral compactness or fine-structure peakedness: high values indicate narrow, resonant peaks characteristic of periodic vowels, while low values indicate the broad, diffuse energy distribution of voiceless fricative turbulence and stop bursts.

\textbf{Effect size and statistical testing}:  All comparisons between synthesised and natural speech are quantified using Cliff's Delta, a non-parametric effect size measure with a range of [\num{-1}, \num{1}]. Statistical significance is assessed using Dunn's post-hoc test with Benjamini-Hochberg correction for multiple comparisons; only effects significant at $p$ $<$ 0.001 are plotted.

\section{Results}\label{sec:results}

\begin{figure*}[t]
  \centering
  \includegraphics[width=\textwidth]{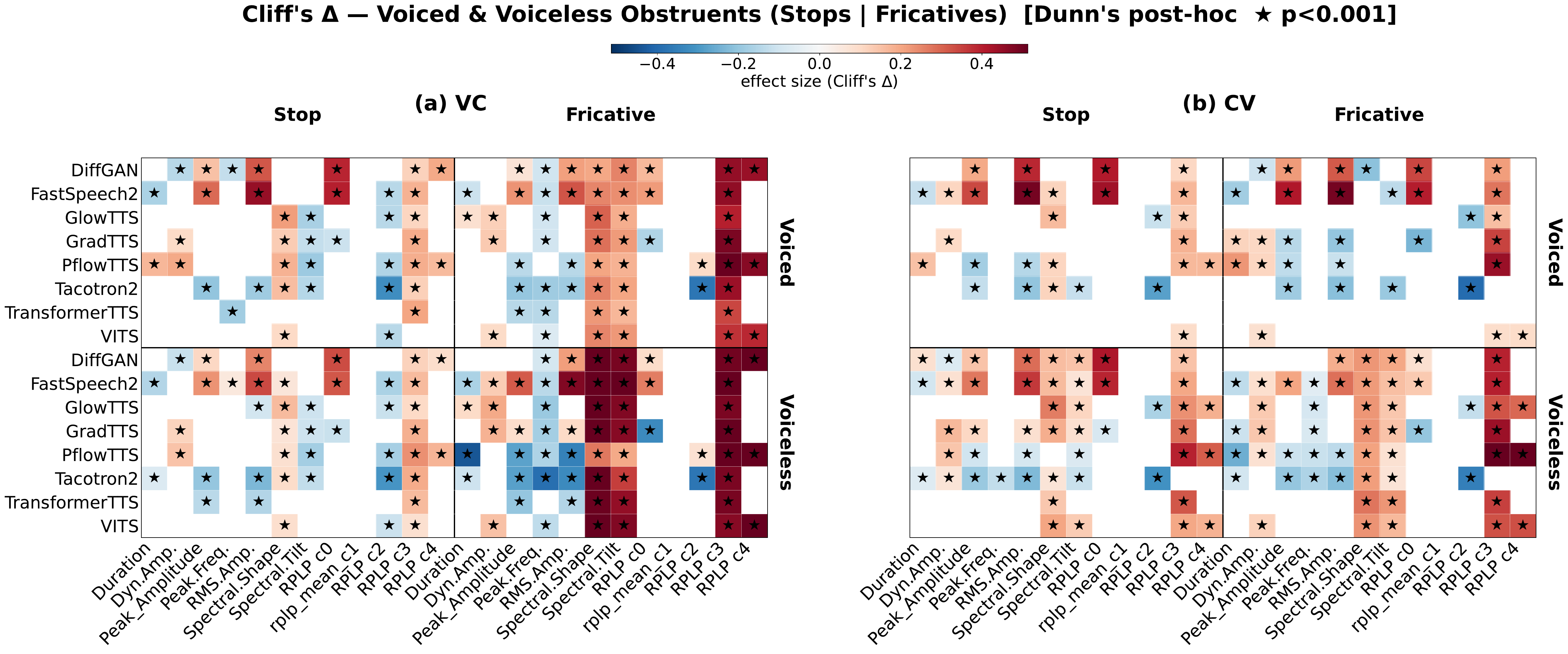}
  \caption{Cliff's $\Delta$ effect sizes for voiced and voiceless obstruents
    (stops and fricatives) across all eight TTS systems.
    Columns correspond to phoneme class (Stop / Fricative) and voicing (Voiced / Voiceless); rows correspond to TTS system.
    Colour encodes effect size: red indicates positive deviation
    (synthesised $>$ natural), blue indicates negative deviation
    (synthesised $<$ natural), and white indicates no significant difference.
    Cells marked with $*$ are significant at $p < 0.001$.
    Left panel:~(a) VC context; right panel:~(b) CV context.}
  \label{fig:heatmap}
\end{figure*}

Figure~\ref{fig:heatmap} shows Cliff’s Delta effect sizes as a heatmap for eight TTS systems in CV and VC contexts, covering voiced and voiceless stops and fricatives. Red indicates higher feature values in synthetic speech than in human natural speech, blue lower values, and white no meaningful difference; starred cells denote significance at  $p$ $<$ 0.001. Peak Amplitude and RPLP C$_1$ showed no significant deviations and are omitted from further discussion. Affricates were excluded due to limited tokens. A clear contrast appears between voiced and voiceless segments: voiced rows show minimal effects, while voiceless rows display dense, significant deviations across systems and features, indicating that the observed errors are pronounced in voiceless obstruents rather than obstruents in general.

\subsection{Do Deviations Persist Across Architectures?}

The most consistent finding is a positive deviation in \textbf{Spectral Shape}, visible as a solid block of red starred cells spanning all eight systems in the voiceless
fricative columns of both panels. Every system concentrates too much energy in lower frequencies relative to natural fricatives. For voiceless stops the same columns are
red but lighter in intensity, reflecting smaller but still significant effect sizes (fricative VC: $d = 0.270$--$0.574$; fricative CV: $d = 0.228$--$0.275$; stop CV:
$d = 0.112$--$0.261$).

The cepstral analysis reinforces this through \textbf{RPLP C$_3$} (Spectral Compactness), which is the most uniformly saturated column in the voiceless rows of both panels.
Natural voiceless obstruents are characterised by broad, flat-spectrum turbulence noise; a positive C$_3$ deviation means the synthesised segment is spectrally too compact,
resembling a resonant structure more than aperiodic noise. All eight systems show deep red starred C$_3$ cells for voiceless fricatives and stops in both CV and VC (fricative
VC: $d = 0.481$--$0.629$; fricative CV: $d = 0.322$--$0.614$; stop VC: $d = 0.161$--$0.235$; stop CV: $d = 0.145$--$0.396$). The sole exception is Tacotron2, whose C$_3$ cell for voiceless fricatives in CV is noticeably pale relative
to all other systems   addressed in Section~\ref{sec:arch}. The convergence of Spectral Shape and RPLP C$_3$ across both feature sets confirms that spectral over-concentration is a cross-paradigm property of all the neural TTS chosen for the study, not an artifact of any individual system or measurement representation. \textbf{Spectral Tilt} follows the same direction for fricatives, as seen in the Spectral Tilt column in the voiceless fricative block, which is uniformly red and
starred across both panels (CV: $d = 0.070$--$0.227$; VC: $d = 0.194$--$0.483$). For stops the Spectral Tilt pattern is context-dependent and is discussed below.

\subsection{Does Context Reveal Asymmetric Patterns?}
\label{sec:context}

Comparing panels (a) and (b) directly reveals that VC and CV produce qualitatively different error profiles. For \textbf{voiceless fricatives}, the Spectral Shape and
Spectral Tilt columns are visibly more saturated in the VC panel than in CV, suggesting that the vowel-to-fricative transition is harder to model than the reverse. \textbf{Peak
Frequency} shows predominantly negative deviations (blue cells) in the VC fricative block for GlowTTS ($d = -0.248$), PFlowTTS ($d = -0.197$), Tacotron2 ($d = -0.130$), and GradTTS ($d = -0.117$), with weaker effects in CV. A depressed spectral peak in VC is consistent with residual vowel energy bleeding into the fricative region at the boundary.

For \textbf{voiceless stops}, the most informative comparison is the Spectral Tilt column between the two panels. In the \textbf{VC stop} block, Tacotron2, GlowTTS, and PFlowTTS
show blue cells, indicating excess high-frequency energy persisting into the closure phase. In the CV panel, the same systems shift toward red or white, suggestive of 
premature low-frequency energy bleeding in before the vowel onset instead. These are complementary boundary failures in opposite directions, visible as a colour sign-flip
in the Spectral Tilt column between panels. The \textbf{RPLP C$_2$} column for voiceless stops shows a parallel effect: compare the sparse blue in CV to the six blue starred cells in VC, strongly indicative of the vowel-to-stop boundary being spectrally harder to model across most architectures.

\subsection{Are There Architecture-Specific Profiles?}
\label{sec:arch}

Reading across individual system rows in the voiceless blocks reveals three distinct signatures. \textbf{FastSpeech2 and DiffGAN} share a row profile dominated by deep red RMS Amplitude cells for both stops and fricatives in both panels. These are the most saturated
RMS values in the heatmap, meaning their obstruents are consistently too loud (FastSpeech2 VC: $d = 0.474$; DiffGAN VC: $d = 0.216$ for fricatives; FastSpeech2
CV: $d = 0.365$; DiffGAN CV: $d = 0.271$ for stops). Their RPLP C$_0$ stop cells are also among the deepest red in the cepstral block (DiffGAN CV: $d = 0.415$; FastSpeech2 CV: $d = 0.391$). It could be attributed to the fact that FastSpeech2 predicts energy deterministically via its variance adaptor; DiffGAN inherits this prior and its adversarial training does not seem to correct the bias for aperiodic segments.

\textbf{Tacotron2} is the most visually distinctive row. Its RPLP C$_2$ cells for voiceless fricatives are the deepest blue in the heatmap in both panels (VC: $d = -0.373$; CV: $d = -0.199$), reflecting global spectral over-smoothing, consistent behavior considering its MSE reconstruction loss. It is also the only system with a pale C$_3$ cell for voiceless fricatives in CV, suggesting that it models natural broad-band turbulence accurately. The reason is hypothesized to be its smoothing suppressing all spectral detail indiscriminately.

\textbf{GradTTS} has one cell that stands out against the otherwise red fricative block: a deep blue RPLP C$_0$ in the VC panel ($d = -0.326$), indicating strong under-estimation of perceived energy in post-vocalic fricatives. This isolated blue
cell is not shared by the other MAS-based systems and points to a behavior specific to GradTTS's diffusion decoder at the VC boundary.

\textbf{PFlowTTS} shows deep blue RMS Amplitude cells for voiceless fricatives in both panels (VC: $d = -0.435$; CV: $d = -0.218$)   the most negative RMS values in the heatmap, while the other MAS-based systems sit near white for the same
feature. This could suggest that its normalizing flow decoder suppresses fricative energy more aggressively than its alignment strategy alone would predict.

\textbf{VITS}, the only end-to-end system, shows the same dense red C$_3$ and C$_4$ pattern as the two-stage systems, with its C$_4$ VC fricative cell among the deepest red in the cepstral block ($d = 0.652$). Since VITS bypasses the mel-spectrogram entirely, this could indicate that spectral over-concentration is embedded in the learned acoustic representations and is not a vocoder artifact.

\section{Discussion}\label{sec:discussion}

The results converge on four interpretive themes. Since this study does not include a perceptual evaluation, all hypotheses concern acoustic and spectral structure only.

\subsection{Voiceless Obstruents as a Diagnostic Region}

The voiced--voiceless contrast in Figure~\ref{fig:heatmap} is the clearest finding: voiced rows are mostly pale, while voiceless rows are dense and saturated across systems, features, and contexts. Voiceless segments are not merely sensitive regions; they show errors large enough to be detected by structurally different analyses. Spectral Shape and Spectral Tilt from acoustic phonetics and RASTA-PLP C$_3$ from an auditory filterbank converge on the same pattern. This agreement across features indicates a genuine distortion rather than a measurement artifact and supports using voiceless obstruents as a focused benchmark for future TTS evaluation.
\label{sec:region}

\subsection{RASTA-PLP as an Engineering-Oriented Diagnostic}

Prior segmental evaluation has relied on acoustic-phonetic features grounded in speech science. RASTA-PLP introduces a complementary perspective interpretable from an
engineering standpoint, with its auditory frequency warping and temporal band-pass filtering making it sensitive to the rapid spectral transitions at CV and VC boundaries.

The recovery of the same deviation patterns in both feature sets has two practical implications. For \textbf{anti-spoofing}, the spectral compactness captured by C$_3$ can be detected from perceptually weighted features, suggesting voiceless obstruent boundaries are useful cues for identifying synthetic speech. For \textbf{architecture development}, RASTA-PLP provides a perceptually grounded proxy that allows engineers to assess whether architectural changes reduce
aperiodic segment distortion without the cost of a full listening experiment.

\subsection{Context-Dependent Asymmetry at Phonetic Boundaries}

VC and CV contexts produce qualitatively different error profiles, not merely different magnitudes. For \textbf{fricatives}, Spectral Shape and Spectral Tilt deviations are consistently larger in VC than CV, suggesting a transition from a periodic vowel
representation to aperiodic noise is harder than the reverse. A plausible mechanism is that the model's harmonic prior decays slowly past the boundary, bleeding vowel energy into the fricative region. For \textbf{stops}, the asymmetry is qualitative rather than quantitative. In VC, Tacotron2, GlowTTS, and PFlowTTS show negative Spectral Tilt, meaning excess high-frequency
energy persists into the closure phase after the vowel ends. In CV, the same systems shift to positive Spectral Tilt, meaning low-frequency harmonic energy bleeds in prematurely before the vowel onset. The failure at each boundary is the mirror image
of the other: in VC the model cannot cleanly exit the vowel; in CV it cannot cleanly delay it. The C$_2$ pattern reinforces this, broadening from two affected systems in CV to six in VC, confirming that the vowel-to-stop direction is spectrally harder to model across most architectures. Prior work~\cite{bhyravajulla25_ssw} hypothesised that context modulates deviation patterns; these results show the modulation is directional, not merely a difference in magnitude.

\subsection{Architecture- and Alignment-Specific Patterns}

\textbf{FastSpeech2} and \textbf{DiffGAN} both show the deepest red RMS
Amplitude cells for voiceless obstruents. FastSpeech2’s deter-
ministic variance adaptor predicts energy from text features,
which works adequately for voiced segments but produces a
consistent upward bias for aperiodic segments whose energy is
highly variable. DiffGAN inherits this prior from FastSpeech2
and its adversarial training does not correct the bias.
\textbf{Tacotron2} shows the deepest blue C$_2$ cells and is the only
system without a significant positive C$_3$ deviation for voiceless
fricatives in CV. Its MSE loss biases the decoder toward the
mean of a predominantly voiced training distribution, producing a globally smoothed envelope. The absent C$_3$ deviation is
not evidence of accurate aperiodic modelling; over-smoothing
suppresses all spectral detail indiscriminately.

\textbf{MAS-based models (GlowTTS, GradTTS, PFlowTTS)} differ from the energy overestimation of FastSpeech2/DiffGAN and the global smoothing of Tacotron2. MAS enforces hard, non-overlapping boundaries. At VC stop closures all three systems show negative
Spectral Tilt and C$_2$ deviations. The abrupt alignment cut suppresses low-frequency vowel harmonics at the boundary but
leaves high-frequency components unattenuated, producing a
spectrally imbalanced transition. This is a boundary-localised failure rather than a global one, and it is specific to the VC direction where the model must switch from a harmonically rich vowel to silence. Within this group, the decoders introduce additional divergence. GradTTS shows an isolated deep blue C$_0$ for voiceless fricatives in VC not seen in GlowTTS or PFlowTTS, suggesting its diffusion decoder under-estimates energy specifically at this transition. PFlowTTS shows the most negative RMS values for voiceless fricatives across both contexts, indicating its normalising flow decoder suppresses fricative energy beyond what the shared alignment strategy would predict. \textbf{VITS}, despite bypassing the mel-spectrogram entirely, shows the same C$_3$ and C$_4$ over-concentration as two-stage systems. The bias toward formant-like spectral structure appears embedded in the VAE latent space rather than the synthesis pipeline, suggesting voiced-dominant training data biases latent representations toward periodic structure independently of synthesis.

\section{Conclusion}\label{sec:conclusion}

This study evaluated voiceless obstruents across eight neural TTS systems at CV and VC boundaries using acoustic-phonetic and RASTA-PLP features. All models showed consistent positive deviations in Spectral Shape, Spectral Tilt, and C$_3$, indicating spectral over-concentration in aperiodic segments is architecture-agnostic. Errors differed across contexts: fricatives showed larger deviations at VC positions, likely due to residual harmonic energy from the preceding vowel, while stops showed opposite patterns across boundaries. FastSpeech2 and DiffGAN overestimated amplitude; Tacotron2 showed strong spectral smoothing; GlowTTS and PFlowTTS exhibited boundary-localized spectral imbalance from MAS enforcing hard alignment cuts leaving high-frequency components unattenuated at VC transitions. VITS showed the same C$_3$ over-concentration despite generating waveforms directly, suggesting the issue stems from latent representations. These findings indicate better handling of sharp spectral transitions is needed in current TTS models.

\section{Generative AI Use Disclosure}
Generative AI was used only for editing and polishing manuscripts, and not for producing a significant part of the manuscript. AI-assisted tools were also used to aid the development of some analysis scripts.

\bibliographystyle{IEEEtran}
\bibliography{mybib}

\end{document}